\pdfoutput=1
\documentclass[11pt]{article}

\usepackage{arxiv}

\usepackage{times}
\usepackage{latexsym}

\usepackage{graphicx}
\usepackage{algorithm}
\usepackage{algpseudocode}

\usepackage{custom}

\usepackage[T1]{fontenc}
\usepackage[utf8]{inputenc}

\usepackage{microtype}

\pgfplotsset{compat=1.17}

\title{InvBERT: Reconstructing Text from Contextualized Word Embeddings by inverting the BERT pipeline}

\author{ 
    Kai Kugler\\
	Trier University \\
	\texttt{kuglerk@uni-trier.de}
	\And
	Simon Münker\\
	Trier University
	\And
	Johannes Höhmann \\
	Trier University
	\And
	Achim Rettinger \\
	Trier University \\
	\texttt{rettinger@uni-trier.de}
}

\begin{document}
\maketitle

\begin{abstract}
Digital Humanities and Computational Literary Studies apply text mining methods to investigate literature. Such automated approaches enable quantitative studies on large corpora which would not be feasible by manual inspection alone. However, due to copyright restrictions, the availability of relevant digitized literary works is limited. Derived Text Formats (DTFs) have been proposed as a solution. Here, textual materials are transformed in such a way that copyright-critical features are removed, but that the use of certain analytical methods remains possible.
Contextualized word embeddings produced by transformer-encoders (like BERT) are promising candidates for DTFs because they allow for state-of-the-art performance on various analytical tasks and, at first sight, do not disclose the original text.
% We argue that one promising candidate for DTFs are contextualized word embeddings produced by transformer-encoders, like BERT. They allow for state-of-the-art performance on various analytical tasks, potentially without disclosing the original text.
However, in this paper we demonstrate that under certain conditions the reconstruction of the original copyrighted text becomes feasible and its publication in the form of contextualized token representations is not safe. 
Our attempts to invert BERT suggest, that publishing the encoder as a black box together with the contextualized embeddings is critical, since it allows to generate data to train a decoder with a reconstruction accuracy sufficient to violate copyright laws. 
% Our attempts to invert BERT suggest, that publishing parts of the encoder is critical, since it allows to generate data to train a decoder with a reconstruction accuracy sufficient to violate copyright laws. Just publishing the contextualized embeddings however, might be safe.
\end{abstract}
\section{Introduction}

Due to copyright laws the availability of text material, specifically literary works
%, for scientific analyses 
is quite limited. Depending on national law there might be some degree of freedom to use protected texts for scientific studies and give reviewers access to them, but in most cases they still can't be published fully, making it hard for the research community to reproduce or build on scientific findings.

This is a fundamental issue for research fields like Digital Humanities (DH) and Computational Literary Studies (CLS), but applies to any analysis of text documents that cannot be made available due to privacy reasons, copyright restrictions or business interests. This, for instance, makes it hard for digital libraries to offer their core service, which is the best possible access to their content. While they provide creative compromise solutions, like \emph{data capsules} or \emph{web-based analysis tools}\footnote{see \url{https://www.hathitrust.org/htrc\_access\_use}}, such access is always limited and complicates subsequent use and reproducibility.

\begin{figure}
\centering
  \includegraphics[width=0.6\linewidth]{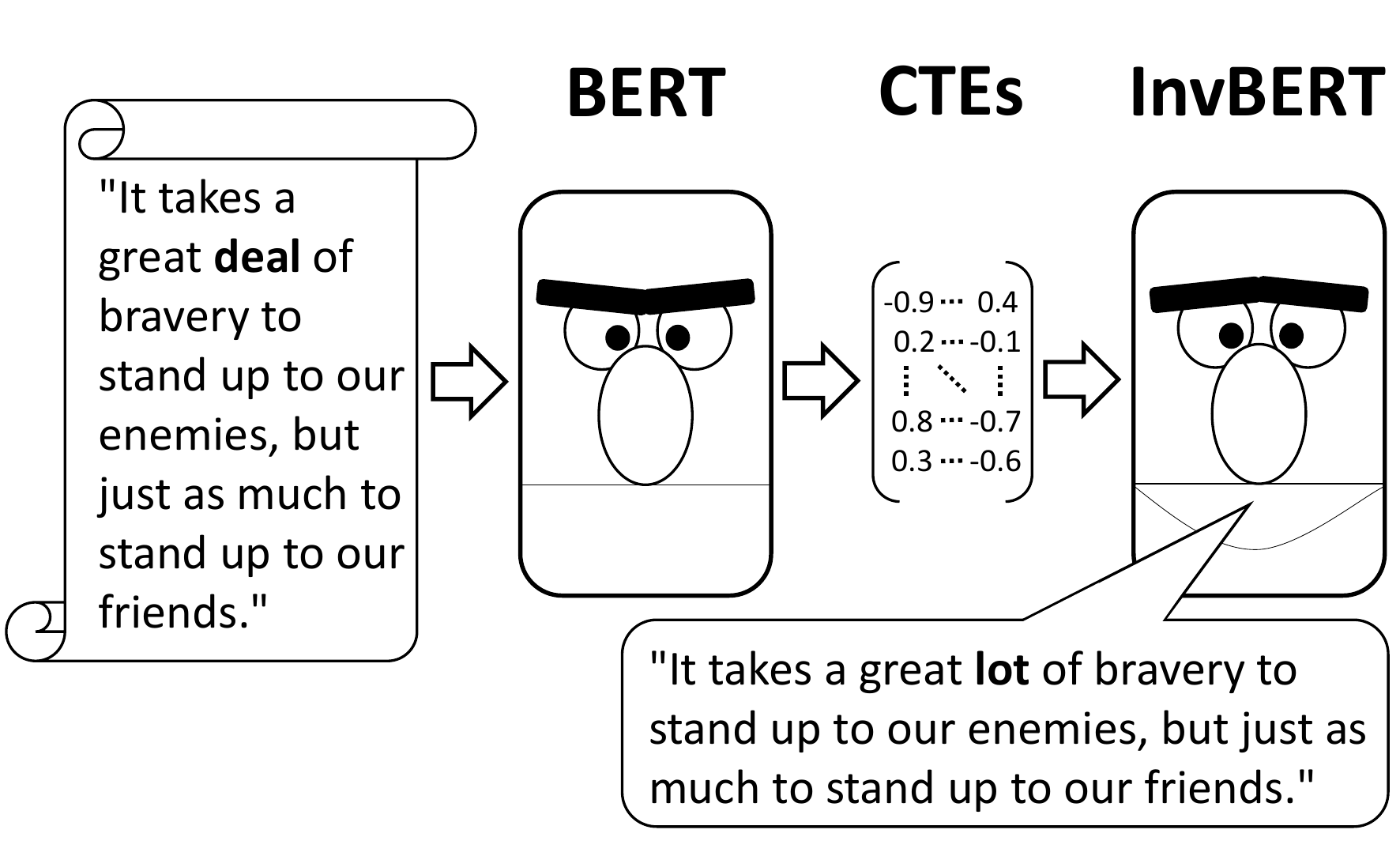}
  \caption{Sample text reconstruction to a Harry Potter quote \cite{rowling1998stone} by inverting BERT.}
  \label{fig:InvBertSimplePipe}
\end{figure}

As a consequence, there have been attempts to find a representation formalism which retains as much linguistic information as possible while not disclosing the original text fully. Such text representations have been referred to as Derived Text Formats (DTFs) \cite{schoch2020abgeleitete}.  
While such DTFs are always a compromise between the degree of obfuscation (non-reconstructibility) and degree of analyzability (retained information), there are DTFs with clear advantages over others. 
%In the end, they should always be more informative than not publishing the documents at all.

We investigate if Contextualized Token Embeddings (CTE), like the ones obtained from a transformer encoder stack trained on a self-supervised masked language modeling (MLM) task \cite{devlin2018bert}, are a promising candidate for DTFs. On the one hand, they are the state-of-the-art text representation for most Natural Language Understanding tasks \cite{wang2018glue,wang2019superglue}, including tasks relevant to DH and CLS, like text classification, sentiment analysis, authorship attribution or text re-use \cite{schoch2020abgeleitete2}. On the other hand, it appears difficult to reconstruct the original text, just from its CTEs.
Thus, we pose the following research question: 
\begin{quote}
In which scenarios can protected text documents be released publicly if encoded as contextualized embeddings since the original content cannot be reconstructed to an extent that violates copyright laws?
\end{quote}

After presenting related work (Sec.~\ref{sec:relatedWork}) we will first formalize different reconstruction scenarios, which
%components used in a text encoder pipeline in order to describe potential application scenarios in DH or CLS. This 
allow us to define potential lines of attack that aim at reconstructing the original text (Sec.~\ref{sec:setup}). Next, we will discuss the feasibility of each line of attack. In Sec.~\ref{sec:empiricism} we focus on the most promising lines of attack by evaluating their feasibility empirically (Sec.~\ref{sec:quantEval}), before concluding in Sec.~\ref{sec:conclusion}.
\section{Related Work}
\label{sec:relatedWork}

First, we look at the very recent field of DTFs, before presenting existing work on text reconstruction beyond copyright protected texts.

\subsection{Derived Text Formats}

DTFs, like n-grams or term-document matrices are an important tool to the Computational Linguistics and Digital Humanities, since they allow the application of quantitative methods to their research objects. However, they have another important advantage: If the publication of an original text is prohibited, DTFs may still enable reproducibility of research \cite{schoch2020abgeleitete2,schoch2020abgeleitete}.
This is especially important for CLS, where there is only a small ``window of opportunity'' of available manuscripts from the year 1800 to 1920 due to technical issues on the lower and copyright restrictions on the upper boundary. Since this is of permanent concern and an obstacle to free research, tools to widen this window are of great importance to the field. 
Other approaches to tackle this issue, like granting access to protected texts in a closed room setting, come with their own major drawbacks and still do not enable an unhindered exchange of scientific findings. Therefore, in most cases, DTFs like term-document matrices are the best solution available. The aim of these formats is to retain as much information as possible, while minimizing reconstructibility. In reality, however, the latter most often is achieved by compromising on the former. This leads to the variety of feasible analytical down-stream tasks being narrowed.
A format that preserves a noticeable amount of information and is already used as a DTF are word embeddings like Word2Vec \cite{mikolov2013efficient} or GloVe \cite{pennington2014glove}. However, similar to term-document matrices they can only be applied to document-level tasks. Otherwise, there remains considerable doubt regarding their resilience against reconstruction attempts. 
A promising attempt to alleviate that is by using contextualized word - or more precise token - embeddings (CTEs) generated by pretrained language models instead, since the search space to identify a token grows exponentially with the length of the sequence containing it. Additionally, these embeddings carry even more information and achieve SOTA results on various down-stream tasks. 

\subsection{Reconstruction of Information from Contextualized Embeddings}

Recently, attention was drawn to privacy and security concerns regarding large language models due to prominent voices in ethics in AI \cite{bender2021dangers}, as well as a collaboratory publication of the industry giants Google, OpenAI and Apple \cite{carlini2020extracting}. In the latter, the authors demonstrated, that these models memorize training data to such an extend, that it is not only possible to test whether the training data contained a given sequence (membership inference, \cite{shokri2017membership}), but also to directly query samples from it (training data extraction). Other recent research supports these findings and agrees, that this problem is not simply caused by overfitting \cite{song2019auditing,Thomas2020InvestigatingTI}.
Gigantic language models like GPT\nobreakdash-3 \cite{brown2020language} or T5 \cite{raffel2019exploring} were trained on almost the entirety of the available web, which poses a special concern, since sensible information like social security numbers is unintentionally being included. Hence, a majority of the literature focuses on retrieving information about the training data. 
However, we argue that such attacks are less successful in the case of literary works, since a) the goal in this scenario would usually be the reconstruction of a specific work, and b) the attacks are not suited to recover more than isolated sequences. 

A third prominent type of attack which can be performed quite effectively and reveals some information about training data is attribute inference \cite{melis2019exploiting,song2020information,Mahloujifar2021}. It is also of little relevance, since it aims to infer information like authorship from the embeddings, which is non-confidential in a DTF setting anyways. More so, authorship attribution is actually a relevant field of research in the DHs.

The main threat regarding CTEs as DTFs are embedding inversion attacks, where the goal is the reconstruction of the original textual work they represent. However, research on this topic is still limited and most paper focus on privacy. Therefore, very few go beyond retrieval of isolated sensitive information. E.g. \cite{pan2020privacy} showed, that it is possible  to use pattern-recognition and key-word-inference techniques to identify content with fixed format (e.g. birth dates) or specific keywords (e.g. disease sites) with varying degree of success (up to 62\% and above 75\% avg. precision respectively). However, this is easier and the search space smaller, than reconstructing full sequences drawn from the whole vocabulary.

To the best of our knowledge, retrieval of the full original text is covered only by \cite{song2020information}. Using an RNN with multi-set prediction loss in a setting with access to the encoding model as a black-box, they were able to achieve an in-domain F1 score of 59.76 on BERT embeddings. However, since privacy was their concern, they did not consider word ordering in their evaluation, which is crucial when dealing with literary works. Therefore, and since they failed to improve on their results using a white-box approach as well, we believe that the security of the usage of CTEs as DTFs still remains an unanswered question.

When dealing with partial-white- or black-box scenarios, a final type of attack should be kept in mind: Inferences about the model itself. Even though not the goal here, successful model extraction attacks \cite{krishna2019thieves} may transform a black-box situation into a white-box case. However, critical information can even be revealed by fairly easy procedures like model fingerprinting. This was showcased on eight SOTA models by \cite{song2020information}, who were able to identify the model based on a respective embedding with 100\% accuracy. 
\section{Reconstruction Task and Attack Vectors}
\label{sec:setup}

This paper is not about improving or applying transformers, but inverting them. To introduce a reconstruction model (cmp. \cite{Rigaki2020}) we first describe scenarios for possible attacks. Then, we lay out different attack vectors based on the scenarios. 

\subsection{Reconstruction Scenarios}

Formally, the reconstruction scenarios can be defined as follows:

\begin{description}
\item[Given:] Contextualized token embeddings CTEs of a copyright protected literary text\footnote{Typically a book, containing literary works, like poetry, prose or drama.} document $W$ are made available in every scenario. Depending on the scenario additional information is available:

\textbf{WB - White Box Scenario:} The most flexible scenario is given if the encoder $enc()$, including the neural network's architecture and learned parameters, and tokenizer $tok()$ is made openly available in addition to the CTEs. Then, analytical experiments can be conducted by DH researchers that require to adapt/optimize the encoder $enc()$ and/or the tokenizer $tok()$. 

\textbf{BB - Black Box Scenario:} A scenario with little flexibility from the perspective of a DH researcher is given, when the tokenizer $tok()$ and the encoder $enc()$ are made available as one single opaque function and are only accessible for generating mappings from $W$ to CTE. A similar scenario arises if ground truth training data is available (i.e., aligned pairs of $W$s to CTEs are given). Then the researcher is still able to label his own training data and use it to optimize $enc()$ or embed other data not yet available as CTEs for analysis. 
However, if provided as a service, the number of queries allowed to be sent to $enc()$ might be limited up to a point where the model is not released at all\footnote{The latter scenario is not considered BB anymore and not covered in this paper (see Sec.~\ref{subsec:futureWork}).}. Then, existing implementations can be reused in order to perform a standard analytical task if the respective task-specific top layer function is also provided. 
%Still, even without the ability to query $enc()$ training data can be obtained if original versions of texts are obtained and aligned with the CTEs. 
Note, that BB can be turned into WB by successful model extraction attacks.

\textbf{GB - Gray Box Scenario:} If the encoder-transformer pipeline $tok()$ and $enc()$ used for generating CTEs is available to some degree (e.g., the tokenizer is given) we refer to it as a Gray Box (GB) scenario.

%Additionally, if the bidirectional-encoder-transformer pipeline $tok()$ and $enc()$ used for generating CTEs is available to some degree we refer to it as a Grey Box (GB) scenario, if the full architecture and all parameters are given a White Box (WB) scenario. If $tok()$ and $enc()$ are one single opaque function and only accessible for generating mappings from $W$ to CTE, or if only ground truth training data is available (i.e., an aligned subset of $W$ to CTE is given) we refer to it as a Black Box scenario (BB).

\item[Searched:] A function or algorithm $inv(CTE)=\hat{W}$ that inverts the model pipeline or approximate its inverse and outputs reconstructed text $\hat{W}$ from CTEs.
\end{description}

\subsection{Inversion Attacks}

We consider three lines of attack:
\begin{description}
\item[Inverting Functions:] Inverting $enc()$ and $tok()$ using calculus requires to find a closed-form expression for $tok^{-1}()$ and $enc^{-1}()$. 
Since this requires knowledge of the parameters of the encoder pipeline, this is only applicable to a WB scenario. Even then, this approach would only be feasible if all functions in question are invertible which is not the case for BERT-like transformer encoder stacks.

\item[Exhaustive Search:] Sentence-by-sentence combinatorial testing of generated inputs to ``guess'' the contextualized token embeddings would be applicable to WB, GB and BB, as long as an unlimited number of queries to $enc()$ is allowed. However, combinatorial explosion renders this approach infeasible: A sentence of 15 tokens
results in $18\cdot10^{66}$ possible combinations, assuming a vocabulary size of 30,522 different tokens, like in the case of BERT\textsubscript{BASE}.

\item[Machine Learning:] Learning an approximation of $tok^{-1}(enc^{-1}())$ can be attempted as soon as training samples are available or can be generated. 
%If the encoder $enc()$ is available and it's use is not in any way limited, an unlimited number of training pairs can be generated.
We assume that an attack is more likely to be successful if components of the embedding generating pipeline are accessible, because in a GB scenario the components can be estimated separately, reducing the complexity compared to an end-to-end BB scenario.
\end{description}

% tbc display flowchart for meeting
%
% tikz flow chart:
%
\tikzstyle{io} = [rectangle, rounded corners, thick, minimum width=1.6cm, minimum height=0.75cm,text centered, draw=black, fill=red!30]
\tikzstyle{process} = [rectangle, thick, minimum width=2.0cm, minimum height=0.75cm, text centered, draw=black, fill=orange!30]
\tikzstyle{ext} = [fill=yellow!30]
\tikzstyle{given} = [densely dashed, fill=yellow!12, inner sep=6pt]
\tikzstyle{decision} = [diamond, double, minimum width=1.5cm, minimum height=0.75cm, text centered, draw=black, fill=green!30]
\tikzstyle{arrow} = [thick,->,>=stealth]

\begin{figure*}[ht]
    \centering
    \begin{subfigure}[b]{0.45\textwidth}
        \centering
        \begin{tikzpicture}[node distance=1.5cm]
            \tikzstyle{every node}=[font=\footnotesize]
            %
            % Nodes:
            %
            \node (txt) [io] {Input Text};

            \node (enc) [process, ext, below of=txt] {\textbf{BERT Pipeline}};

            \node (tok) [io, below left of=enc, xshift=-1.2cm] {Tokens};
            \node (idx) [io, below of=enc, yshift=0.45cm] {Vocab IDs};
            \node (cte) [io, below right of=enc, xshift=1.2cm] {CTEs};

            \node (mdl) [process, below of=cte] {MLP Classifier};

            \node (prd) [io, below of=mdl, yshift=-0.4cm] {Predicted IDs};

            \node (lss) [decision, below of=idx, yshift=-0.2cm] {Loss};

            \node (met) [decision, below of=tok, yshift=-0.2cm] {Metric};

            \node (dec) [process, ext, below of=lss, yshift=-1.6cm] {\textbf{BERT Pipeline}};

            \node (ptx) [io, below of=met, yshift=-0.2cm] {Reconstructed Tokens};

             % draw rectangle around given components
            \begin{scope}[on background layer]
                \node (giv) [given, draw, fit=(txt)(idx)(cte)(tok)] {};
                \node (giv) [given, draw, fit=(dec)] {};
            \end{scope}

            % -----------------------------
            %
            % Path
            %

            \draw[arrow] ++(-2cm,0) -- (txt);
            \draw [arrow] (txt) -- (enc);

            \draw [arrow] (enc) -| (tok);
            \draw [arrow] (enc) -- (idx);
            \draw [arrow] (enc) -| (cte);

            \draw [arrow] (cte) -- (mdl);
            \draw [arrow] (idx) -- (lss);

            \draw [arrow] (mdl) -- (prd);

            \draw [arrow] (prd) -| (lss);
            \draw [arrow] (prd) -| (dec);

            \draw [arrow] (dec) -| (ptx);

            \draw [arrow] (tok) -- (met);
            \draw [arrow] (ptx) -- (met);
        \end{tikzpicture}
        \caption{Our InvBERT Classify approach retrieves tokens, IDs, CTEs from the encoder (BERT Pipeline) and utilizes a multi-layer classifier to predict IDs. We use the identical encoder to reconstruct the original token/text.}
        \label{fig:flowchart:linear}
    \end{subfigure}
    \hfill
    \begin{subfigure}[b]{0.45\textwidth}
        \centering
        \begin{tikzpicture}[node distance=1.5cm]
            \tikzstyle{every node}=[font=\footnotesize]
            %
            % Nodes:
            %
            \node (txt) [io] {Input Text};

            \node (tok) [process, below left of=txt, xshift=-1.6cm, yshift=-0.4cm] {\textbf{Tokenizer}};
            \node (enc) [process, ext, below of=txt, yshift=0.0cm] {BERT Pipeline};

            \node (tks) [io, below left of=tok, xshift=-1.2cm, yshift=-0.45cm] {Tokens};
            \node (idx) [io, below of=tok, yshift=0cm] {Vocab IDs};
            \node (cte) [io, below of=enc, yshift=0.05cm] {CTEs};

            \node (mdl) [process, below of=cte] {Transformer Decoder};

            \node (prd) [io, below of=mdl, yshift=0.0cm] {Predicted IDs};

            \node (lss) [decision, below of=idx, yshift=-0.2cm] {Loss};

            \node (met) [decision, below of=tks, yshift=-0.2cm] {Metric};

            \node (dec) [process, below of=lss, yshift=-1.2cm] {\textbf{Tokenizer}};

            \node (ptx) [io, below of=met, yshift=-0.2cm] {Reconstructed Tokens};

            \begin{scope}[on background layer]
                \node (giv) [given, draw, fit=(txt)(enc)(cte)] {};
            \end{scope}

            % -----------------------------
            %
            % Path
            %

            \draw[arrow] ++(-2cm,0) -- (txt);
            \draw [arrow] (txt) -- (enc);
            \draw [arrow] (txt) -- (tok);

            \draw [arrow] (tok) -| (tks);
            \draw [arrow] (tok) -- (idx);
            \draw [arrow] (enc) -- (cte);

            \draw [arrow] (cte) -- (mdl);
            \draw [arrow] (idx) -- (lss);

            \draw [arrow] (mdl) -- (prd);

            \draw [arrow] (prd) -| (lss);
            \draw [arrow] (prd) -| (dec);

            \draw [arrow] (dec) -| (ptx);

            \draw [arrow] (tks) -- (met);
            \draw [arrow] (ptx) -- (met);
        \end{tikzpicture}
        \caption{In InvBERT Seq2seq we train a custom tokenizer (BytePair) and utilize only the given CTEs (BERT Pipeline) to sequentially predict token IDs utilizing a Transformer Decoder Structure. Here we use our tokenizer to reconstruct the original token/text.}
        \label{fig:flowchart:seq2seq}
    \end{subfigure}
    \caption{Flowchart for each approach. Givens are enclosed in a dotted yellow area and  attack-specific modules to be estimated are filled with orange. Data objects are highlighted in red, while green represent the evaluation/objective function.}
    \label{fig:flowchart}
\end{figure*}
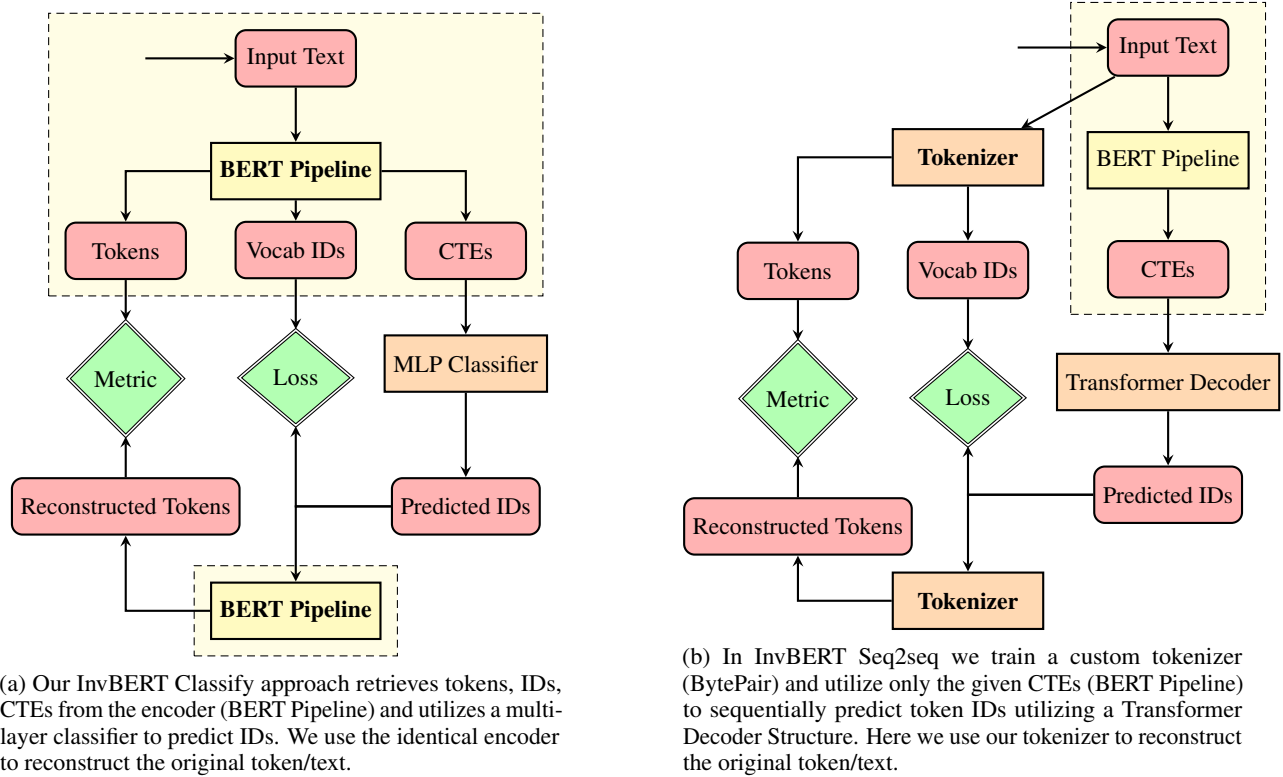

Since a successful BB attack equally works in a GB scenario and a successful GB attack works in a WB scenario we restrict our empirical investigation to two machine learning based attacks, one for a GB, where $tok()$ is given and one for the BB scenario. We call our GB attack \emph{InvBert Classify} and our BB attack \emph{InvBert Seq2Seq}.
Both models are detailed in Fig.~\ref{fig:flowchart} and described in the next section.
\section{Experimental Design}
\label{sec:empiricism}
In this Section, we describe two attack models, one for a GB and one for a BB scenario, introduced in Section \ref{sec:setup}. First, we introduce and discuss the datasets. Next, we explain both neural network structures and the general attack pipeline. The code and datasets are publicly available as a Github repository.\footnote{Available on GitHub, once the paper is accepted (during review as an anonymous repository): \url{https://anonymous.4open.science/r/invbert-BF31}. The AO3 corpus cannot be made available, for the same copyright reasons discussed in this paper. However, it can be recrawled to replicate our experiments. The Gutenberg corpus is freely downloadable and usable}. 

\subsection{Data}
\begin{table}
    \centering
    \small
    \begin{tabular}{| l | r r | r r |}
        \hline
        & \multicolumn{2}{c|}{\textbf{Train}} 
        & \multicolumn{2}{c|}{\textbf{Eval}}
        \\
        \textbf{Name} & \textbf{Size} & \textbf{Samples} & \textbf{Size} & \textbf{Samples}
        \\\hline
        Ao3 Action  & 391\,MB  & 5903k     & 10\,MB  &   146k \\
        Ao3 Drama   & 319\,MB  & 4854k     & 9\,MB   &   121k \\
        Ao3 Fluff   & 343\,MB  & 5251k     & 9\,MB   &   131k \\
        Gutenberg   & 270\,MB  & 2728k     & 7\,MB   &   68k \\
        \hline
    \end{tabular}
    \caption{Size and number of contained training samples of the collected data sets.}
    \label{fig:table:data}
\end{table}
We chose two openly available text corpora which resemble protected work and contain a sufficient amount of text documents to generate suitable data sets.
%This does not mean that the texts are necessarily copyright free or freely distributable, but they are chosen to be similar to the texts that are to be distributed in DTF and that our experiments are replicable.
Table \ref{fig:table:data} shows the exact size and number of samples of each subset.

First, we scraped the \emph{Archive of Our Own (AO3)}\footnote{\url{https://archiveofourown.org}}, an openly available fanfiction repository, using a modified version of AO3Scraper\footnote{\url{https://github.com/radiolarian/AO3Scraper}}. During the prepossessing, we filtered out mature, extreme, and non-general audience content using the given tags. We split the AO3 data into the following three topics based on the ten most common tags: \emph{Action}, \emph{Drama} and \emph{Fluff}\footnote{"Feel good" fan fiction designed to be happy, and nothing else, according to \url{https://en.wikipedia.org/wiki/Fan\_fiction}}. 

As fanfiction mostly resembles contemporary literature, we gathered a fourth dataset from Project Gutenberg\footnote{\url{https://www.gutenberg.org/}}, a non-commercial platform but with a focus on archiving and distributing historical literature, including western novels, poetry, short stories, and drama. Consequently, our Gutenberg train/eval set contains a mix of different genres in contrast to the AO3 datasets. Gutenberg's content is sorted by \emph{bookshelves}, we have selected prose genres in Modern English (Classics, Fiction, Adventure etc.) not removing any metadata. 

% commonly written in Middle English or Early Modern English, being substantially different to the data used for the pre-training of BERT. 
%We suspect those differences also affect the results in our experiments.

\subsection{Models \& Pipelines}
In Sec.~\ref{sec:setup} we argued that machine learning models are promising candidates for inversion attacks. We propose two models, one for a GB and one for a BB scenario: 

\begin{description}
    \item[InvBERT Classify (GB):] Here, we have access to the CTEs and the tokenizer $tok()$. As the tokenizer is a look-up table, which can be queried from both directions, the inverse $tok^{-1}()$ to $tok()$ is also provided, effectively simplifying the problem of finding an approximation of the inverse $tok^{-1}(enc^{-1}())$ of the whole pipeline to just $enc^{-1}()$. We train a multi-layer perceptron to predict the vocabulary IDs given CTEs. As we use the given tokenizer, CTEs and IDs have a one-to-one mapping, and our attack boils down to a high-dimensional token classification task. 
    \item[InvBERT Seq2Seq (BB):] Here, we only have access to the CTEs. Without the tokenizer, we lose the one-to-one mapping and cannot infer the token CTE ratio. Thus, we have to train a custom tokenizer and optimize a transformer decoder structure to predict our sequence of custom input IDs. The decoder utilizes complete sentence CTEs as generator memory and predicts each token ID sequentially.
\end{description}

We use the \emph{Hugging Face API}\footnote{\url{https://huggingface.co}} to construct a batch-enabled BERT Pipeline capable of encoding plain text into CTEs and decoding (sub-) token IDs into words. All parameters inside the pipeline are disabled for gradient optimization. Our models and the training/evaluation routine are based on \emph{PyTorch modules}\footnote{\url{https://pytorch.org}}. We utilize AdamW as an optimizer and the basic cross-entropy loss. Our model implementations have $\sim 24M$ (InvBert Classify) and $\sim 93M$ (InvBERT Seq2Seq) trainable para\-meters.    

We train on a single Tesla V100-PCIe-32GB GPU and do not perform any hyperparameter optimization. Further, we use in each type of attack the identical hyperparameter settings to ensure the highest possible comparability.\footnote{The parameters used for the experiments can be found in the configuration files of the repository} A training epoch for a model takes up to 8 hours depending on the dataset and type of attack. 

\subsection{Evaluation Metrics}
We evaluate the 3-gram, 4-gram, and sentence precision in addition to the BLEU metric \cite{Papineni2002}. The objective of our model is to reconstruct the given input as closely as possible. BLEU defines our lower bound in terms of precision, as it is based on n-gram precision allowing inaccurate sentences with matching sub-sequences. Since the BLEU metric might be too imprecise to quantify if a reconstruction captures the content of a sentence and style of the author, we preferred to use complete sentence accuracy in our quantitative evaluation. There, we only count perfectly correct reconstructions, resulting in a significantly higher bound in contrast to BLEU.
\section{Empirical Results:}
\label{sec:quantEval}

In this section we will first present our qualitative results, before showing some examples of different reconstruction results.

\subsection{Quantitative Evaluation}
\begin{figure*}
    \centering
    \begin{subfigure}[b]{0.475\textwidth}
        \centering
        \begin{tikzpicture}
            \begin{axis}[
                xlabel={Dataset Size},
                ylabel={Sentence Reconstruction Accuracy},
                symbolic x coords={100\%, 10\%, 1\%, 0.1\%},
                % general styles:
                ymajorgrids=true,
                % legend styles:
                legend cell align = {left},
                %legend pos = nord east,
                % axis:
                ymin=0.0,
                ymax=1.1,
                ybar=2*\pgflinewidth,
                bar width=8pt,
                xtick = data,
                scaled y ticks = false,
                enlarge x limits=0.25,
                ymin=0,
                nodes near coords align={vertical},
                ]

                % action (linear)
                \addplot[ybar, style={colorsave_blue,fill=colorsave_blue,mark=none}]
                    coordinates {(100\%, 0.9931) (10\%, 0.9769) (1\%, 0.7899) (0.1\%, 0.2182)};

                % drama (linear)
                \addplot[ybar, style={colorsave_red,fill=colorsave_red,mark=none}]
                    coordinates {(100\%, 0.9931) (10\%, 0.9723) (1\%, 0.7507) (0.1\%, 0.1825)};

                % fluff (linear)
                \addplot[ybar, style={colorsave_yellow,fill=colorsave_yellow,mark=none}]
                    coordinates {(100\%, 0.9928) (10\%, 0.9746) (1\%, 0.7756) (0.1\%, 0.2172)};

                % gutenberg (linear)
                \addplot[ybar, style={colorsave_darkgreen,fill=colorsave_darkgreen,mark=none}]
                    coordinates {(100\%, 0.9744) (10\%, 0.9110) (1\%, 0.5041) (0.1\%, 0.0421)};

                % avg (linear)
                % 100\% = 0.9931 + 0.9931 + 0.9928 + 0.9744
                % 10\% = 0.9769 + 0.9723 + 0.9746 + 0.9110
                % 1\% =  0.7899 + 0.7507 + 0.7756 + 0.5041
                % 0.1\% = 0.2182 + 0.1825 + 0.2172 + 0.0421
                \addplot[draw=colorsave_blue, thick, smooth, mark=square]
                    coordinates {(100\%, 0.9883) (10\%, 0.9587) (1\%, 0.7050) (0.1\%, 0.165)};

            \end{axis}
        \end{tikzpicture}
        \caption{InvBERT Classify: CTEs and tokenizer object are given.}
        \label{fig:plot:results:classify}
    \end{subfigure}
    \hfill
    \begin{subfigure}[b]{0.475\textwidth}
        \centering
        \begin{tikzpicture}
            \begin{axis}[
                xlabel={Dataset Size},
                symbolic x coords={100\%, 10\%, 1\%, 0.1\%},
                % general styles:
                ymajorgrids=true,
                % yticklabels={,,},
                % legend styles:
                legend cell align = {left},
                % legend pos = nord east,
                % axis:
                ymin=0.0,
                ymax=1.1,
                ybar=2*\pgflinewidth,
                bar width=8pt,
                xtick = data,
                scaled y ticks = false,
                enlarge x limits=0.25,
                ymin=0,
                nodes near coords align={vertical},
                ]

               % action (seq2seq)
                \addplot[ybar, style={colorsave_blue, fill=colorsave_blue, mark=none}]
                    coordinates {(100\%, 0.9174) (10\%, 0.8049) (1\%, 0.4009) (0.1\%, 0.0251)};

                % drama (seq2seq)
                \addplot[ybar, style={colorsave_red, fill=colorsave_red, mark=none}]
                    coordinates {(100\%, 0.9050) (10\%, 0.7813) (1\%, 0.3093) (0.1\%, 0.0206)};

                % fluff (seq2seq)
                \addplot[ybar, style={colorsave_yellow, fill=colorsave_yellow, mark=none}]
                    coordinates {(100\%, 0.9064) (10\%, 0.8007) (1\%, 0.3416) (0.1\%, 0.0243)};

                % gutenberg (seq2seq)
                \addplot[ybar, style={colorsave_darkgreen, fill=colorsave_darkgreen, mark=none}]
                    coordinates {(100\%, 0.7528) (10\%, 0.4917) (1\%, 0.0764) (0.1\%, 0.0070)};

                % avg (seq2seq)
                % 100\% = 0.9174 + 0.9050 + 0.9064 + 0.7528
                % 10\% = 0.8049 + 0.7813 + 0.8007 + 0.4917
                % 1\% = 0.4009 + 0.3093 + 0.3416 + 0.0764
                % 0.1\% = 0.0251 + 0.0206 + 0.0243 + 0.0070
                \addplot[draw=colorsave_blue, thick, smooth, mark=square]
                    coordinates {(100\%, 0.8704) (10\%, 0.7196) (1\%, 0.2820) (0.1\%, 0.0192)};

                \legend{
                    Ao3 Action,
                    Ao3 Drama,
                    Ao3 Fluff,
                    Gutenberg,
                    AVG
                }
            \end{axis}
        \end{tikzpicture}
        \caption{InvBERT Seq2Seq: Only CTEs are given.}
        \label{fig:plot:results:seq2seq}
    \end{subfigure}
    \caption{Both reconstruction approaches compared by their in-domain sentence reconstruction accuracy.}
    \label{fig:plot:results}
\end{figure*}
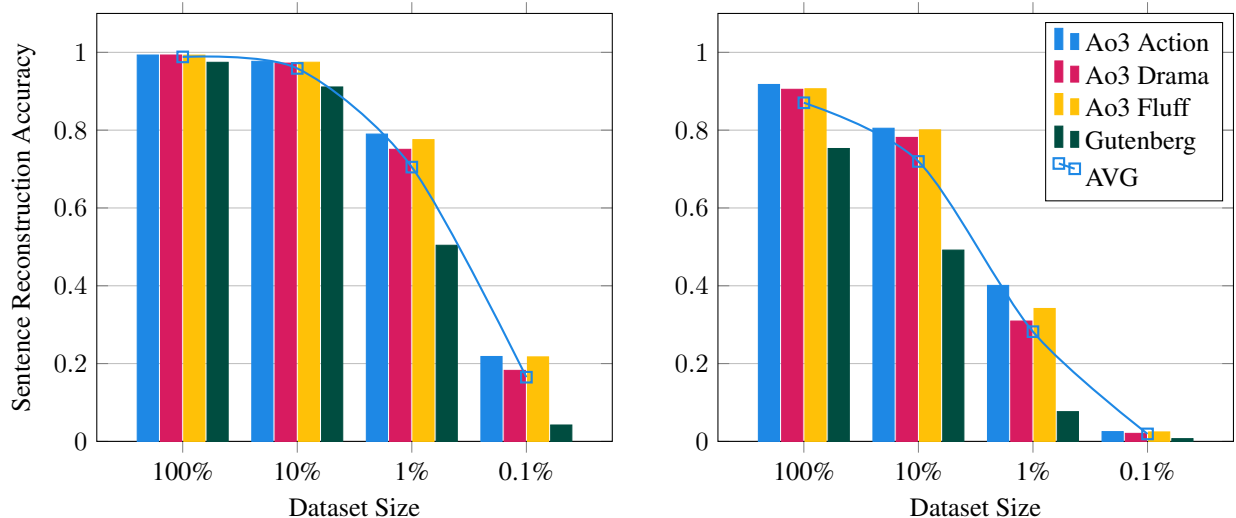

We quantitatively evaluated the trained models in-domain by calculating their sentence accuracy over all samples of their corresponding test set. Equally, we determined out-of-domain performance by repeating the procedure for each model using the respective evaluation data-sets. A condensed representation of our in-domain results is presented in Fig.~\ref{fig:plot:results}, while the full results are included in appendix \ref{sec:appendix:tables}.

The InvBERT Classify model achieves a very high in-domain as well as out-domain sentence reconstruction accuracy when trained on $100\%$ and $10\%$ of the training data-set. Thus, we can reconstruct around $\approx 97\%$ of the original content without errors. Even when just utilizing $1\%$ of the training datasets, our model scores $\approx 65\%$ sentence reconstruction accuracy. This likely still is enough to violate copyright laws since the remaining $35\%$ of sentences get very close to the originals. Only, if we train on $0.1\%$ of the data, the generated text clearly does not resemble the original input.

We observe that the performance on the AO3 datasets, even across genres, is very consistent. The performance considerably drops on the Gutenberg corpus. We assume that the more heterogeneous content in combination with input shuffling during training yields a more challenging data set than our AO3 crawl. In particular, the smaller the train subsets, the smaller the number of samples of a certain genre inside our Gutenberg corpus. Additionally, the Gutenberg corpus contains noise like metadata and unique tokens in the form of title pages and table of contents which we did not clean. The differences are negligible when using $100\%$ or $10\%$ of the training data set, but become clear on $1\%$ or $0.1\%$ train data usage, where the accuracy differs around $20\%$.

The InvBERT Seq2Seq2 model reaches slightly worse results while also being much more sensitive to the training data size and the type of dataset. This is to be expected since this approach utilizes a more complex network architecture that sequentially predicts the reconstruction parts. We attribute the differences to the more complex task and the higher number of trainable parameters.

\subsection{Qualitative Evaluation}
\newcommand{\TRUE}{\textit{\makecell{\centering exact reconstruction}}}
\newcommand{\SYN}[1]{\textcolor{colorsave_yellow}{#1}}
\newcommand{\ERR}[1]{\textcolor{colorsave_red}{#1}}

\begin{table*}[t]
    \centering
    \footnotesize
    \begin{tabularx}{\textwidth}{| l || X | X |}
        \hline
        \B{SRC:}
            &   if you want to know what a man's like, take a good look at how he treats his inferiors, not his equals.
            &   i'll just go down and have some pudding and wait for it all to turn up ... it always does in the end. \\
        \hline
        \multicolumn{3}{c}{} \\
        \multicolumn{3}{c}{\B{InvBERT Classify}} \\
        \hline
        100\%
            &   \TRUE
            &   \TRUE \\\hline
        10\%
            &   \TRUE
            &   \TRUE \\\hline
        1\%
            &   if you want to know what a man's like, take a good look at how he treats his \SYN{subordinates}, not his equals.
            &   \TRUE \\\hline
        0.1\%
            &   if you want to know what a man's like, take a good look at how he treat his \SYN{enemies}, not his \SYN{friends}.
            &   i'll just go down and have some \SYN{dinner} and wait for it all to come up ... it always does in the end. \\
        \hline
        \multicolumn{3}{c}{} \\
        \multicolumn{3}{c}{\B{InvBERT Seq2seq}} \\
        \hline
        100\%
            &  \TRUE
            &  \TRUE \\\hline
        10\%
            &   if you want to know what a man's like, take a good look at how he treats his inferior \ERR{tors}, not his equals.
            & \TRUE \\\hline
        1\%
            &   if you want to know what a man's like, take a good look at \ERR{his partners}, not \ERR{his partners}.
            &   i'll just go down and \ERR{wait for} some \SYN{chocolate} and wait for it all to turn up \ERR{in the end} ... it always does in the end. \\\hline
        0.1\%
            &   if you \ERR{have a little to get a little, but you're a little look at him, not like he're a little look}.
            &   i'll \ERR{go and get up up the rest of the rest , it just just just just have been going to get up}. \\
        \hline
    \end{tabularx}
    \caption{Example of Harry Potter quotes \cite{rowling2006princce} and their predictions. Differences are highlighted: \ERR{red} as error and \SYN{yellow} as false, but semantically acceptable. '\textit{exact reconstruction}' represent identical reproduction.}
    \label{fig:table:quotes}
\end{table*}

To put our previously made assumption about their reconstruction quality to the test, we applied our models to 15 quotes from the Harry Potter book series.\footnote{Retrieved from \url{https://mashable.com/article/best-harry-potter-quotes}} The calculated metrics in Table~\ref{fig:table:quotes} show that the performance on these real-world examples are consistent with the quantitative results on our test data. 

InvBERT Classify completely reconstructs the samples when trained on $100\%$ or $10\%$ of the training dataset. Only when using $1\%$ or $0.1\%$ of the train data, the model predict false but semantically similar content. Contrary, InvBERT Seq2Seq starts to produce substantial errors in its reconstruction while using $10\%$ of the train data, and with less data, the predictions do not resemble a reasonable reconstruction attempt neither on the syntactic nor semantic level.

\subsection{Discussion}

Our exemplary manual evaluation corroborates the results from our quantitative experiments. Both attacks can, if enough data is available, successfully reconstruct the original content. In conclusion, according to our assessment, all scenarios (WB, GB and BB) cannot be considered safe. Even in the ``safest'' BB scenario without a given tokenizer, reconstruction is feasible.

% something about how easy it is to obtain training data

Collecting training data has proven to be very easy, as there are many corpora available digitally that are sufficiently similar to modern English-language texts. The word order information that BERT can extract from this data is apparently sufficient to reconstruct texts from CTEs derived from texts that are not allowed to be published.

Thus, copyright violations are imminent when publishing CTEs as DTFs.
\section{Conclusion and Future Work}
\label{sec:conclusion}

To conclude, we first summarize our contributions and findings, before outlining open research questions.

\subsection{Summary and Conclusion:} 

Derived Text Formats (DTFs) are an important topic in Digital Humanities (DH).  There, the proposed DTFs rely on deleting important information from the text, e.g., by using term-document matrices or paragraph-wise randomising of word orders. In contrast, Contextualized Token Embeddings (CTEs), as produced by modern language models, are superior in retaining syntactic and semantic information of the original documents.
However, the use of CTEs for large-scale publishing of copyright protected works as DTFs is hindered by the risk that the original texts can be reconstructed. 

In this paper we first identify and describe typical scenarios in DH when analyzing text using CTEs is helpful to different degrees. Next, we list potential attacks to recover the original texts. We theoretical and empirically investigate what attack can be applied in which scenario.

Our findings suggest, that if a certain number of training instances (known mappings of sequences of CTEs produced by the encoder to the original sentences) are given or can be obtained it is not save to publish CTEs. Even the safest BB scenario that we covered in this paper is not resistant against reconstruction attacks. Consequently, all GB and WB scenarios are even more vulnerable.

\subsection{Future Work:} 
\label{subsec:futureWork}

While researchers from the area of DH have to judge the usefulness of CTEs as DTFs, finding a copyright protected way of publishing content is also relevant for the field of Natural Language Processing (NLP) in general. There, CTEs have only been investigated in regards to privacy risks, but not copyright protection. After all, the problem of reproducibility of scientific results from restricted corpora is not limited to the DHs.

The focus of this paper is to define the task of reconstructing text from CTEs of literary works, however we encourage to establish a novel research niche beyond DH. Accordingly, we only covered the most obvious lines of attack, there are more scenarios that require additional investigation.

Another potential scenario that has not been discussed in this paper is the publication of CTEs without any (means to generate) training data. While this scenario demands future research, it appears not promising for two reasons: First, to be of any value for DH researchers, the bibliographic meta data (author, title, $\ldots$) of the literary work has to be published along with the CTEs. Still, this is limited, since the rich information encoded in CTEs (e.g., compared to a bag-of-words representation) is hard to leverage without more fine-grained information of the structures of the documents (like sentences).
Second, ensuring that no training data can be obtained from a released sequence of CTEs seams only feasible in very special cases. If (parts of) the literary works in the corpus can be obtained in a digitized format through other means, it might be possible to align them with the sequence of CTEs and generate a training set. How sentences can be aligned remains the key research challenge in such a scenario, but as soon as an alignment can be established it becomes an invertible BB scenario.

Also, there is the question of finding a compromise scenario where the complete sequence of CTEs is not published or noise is added, as it has been done with DTFs. Examples are shuffling the sequence, random deletion of a portion of the CTEs, or representation of certain CTEs by linguistic features. What benefits CTEs provide in such scenarios is also a question for future research.

Of interest for future research also are CTEs generated by more modern language models than BERT. Language models still keep growing in size and capabilities and so do the complexity of the CTEs they generate. If this implies a fundamental change to our findings is to be investigated, but we assume that it is rather a matter of scaling the reconstruction model accordingly, than rendering our general approach infeasible. 

Another related issue that we did not discuss, is the suitability of quantitative metrics for measuring copyright violations. Ultimately, it is a legal consideration, if a reconstruction accuracy, e.g., above a certain BLEU-score, violates copyright laws. This is beyond the scope of this paper.

Ultimately, publishers and libraries need to decide if they release DTFs of their inventory. However, based on our findings we advise against it, since it is likely that training samples might be obtained. Still, we believe that more research in NLP is needed to find compromise solutions that balance usefulness while ensuring safety from reconstruction. What contribution CTEs can provide is still an open question.
%The task of the NLP community is to show what options are feasible and safe.
For NLP researchers, this is an exciting challenge, since it requires both, theoretical studies regarding computational complexity, but also empirical experiments with real-word corpora in real-world settings.

\bibliography{invbert}

\begin{thebibliography}{10}

\bibitem{rowling1998stone}
J.~K. Rowling.
\newblock {\em Harry Potter and the Sorcerer's Stone}.
\newblock Scholastic, United States, 1998.

\bibitem{schoch2020abgeleitete}
Christof Sch{\"o}ch, Fr{\'e}d{\'e}ric D{\"o}hl, Achim Rettinger, Evelyn Gius,
  Peer Trilcke, Peter Leinen, Fotis Jannidis, Maria Hinzmann, and J{\"o}rg
  R{\"o}pke.
\newblock Abgeleitete textformate: Prinzip und beispiele.
\newblock {\em RuZ-Recht und Zugang}, 1(2):160--175, 2020.

\bibitem{devlin2018bert}
Jacob Devlin, Ming-Wei Chang, Kenton Lee, and Kristina Toutanova.
\newblock Bert: Pre-training of deep bidirectional transformers for language
  understanding.
\newblock {\em arXiv preprint arXiv:1810.04805}, 2018.

\bibitem{wang2018glue}
Alex Wang, Amanpreet Singh, Julian Michael, Felix Hill, Omer Levy, and
  Samuel~R. Bowman.
\newblock {GLUE}: A multi-task benchmark and analysis platform for natural
  language understanding.
\newblock In {\em International Conference on Learning Representations}, 2019.

\bibitem{wang2019superglue}
Alex Wang, Yada Pruksachatkun, Nikita Nangia, Amanpreet Singh, Julian Michael,
  Felix Hill, Omer Levy, and Samuel~R Bowman.
\newblock Superglue: A stickier benchmark for general-purpose language
  understanding systems.
\newblock {\em arXiv preprint arXiv:1905.00537}, 2019.

\bibitem{schoch2020abgeleitete2}
Christof Sch{\"o}ch, Fr{\'e}d{\'e}ric D{\"o}hl, Achim Rettinger, Evelyn Gius,
  Peer Trilcke, Peter Leinen, Fotis Jannidis, Maria Hinzmann, and J{\"o}rg
  R{\"o}pke.
\newblock Abgeleitete textformate: Text und data mining mit urheberrechtlich
  geschützten textbeständen.
\newblock {\em Zeitschrift f{\"u}r digitale Geisteswissenschaften}, 2020.

\bibitem{mikolov2013efficient}
Tomas Mikolov, Kai Chen, Greg Corrado, and Jeffrey Dean.
\newblock Efficient estimation of word representations in vector space.
\newblock {\em arXiv preprint arXiv:1301.3781}, 2013.

\bibitem{pennington2014glove}
Jeffrey Pennington, Richard Socher, and Christopher~D Manning.
\newblock Glove: Global vectors for word representation.
\newblock In {\em Proceedings of the 2014 conference on empirical methods in
  natural language processing (EMNLP)}, pages 1532--1543, 2014.

\bibitem{bender2021dangers}
Emily~M Bender, Timnit Gebru, Angelina McMillan-Major, and Shmargaret
  Shmitchell.
\newblock On the dangers of stochastic parrots: Can language models be too big?
\newblock In {\em Proceedings of the 2021 ACM Conference on Fairness,
  Accountability, and Transparency}, pages 610--623, 2021.

\bibitem{carlini2020extracting}
Nicholas Carlini, Florian Tramer, Eric Wallace, Matthew Jagielski, Ariel
  Herbert-Voss, Katherine Lee, Adam Roberts, Tom Brown, Dawn Song, Ulfar
  Erlingsson, et~al.
\newblock Extracting training data from large language models.
\newblock {\em arXiv preprint arXiv:2012.07805}, 2020.

\bibitem{shokri2017membership}
Reza Shokri, Marco Stronati, Congzheng Song, and Vitaly Shmatikov.
\newblock Membership inference attacks against machine learning models.
\newblock In {\em 2017 IEEE Symposium on Security and Privacy (SP)}, pages
  3--18. IEEE, 2017.

\bibitem{song2019auditing}
Congzheng Song and Vitaly Shmatikov.
\newblock Auditing data provenance in text-generation models.
\newblock In {\em Proceedings of the 25th ACM SIGKDD International Conference
  on Knowledge Discovery \& Data Mining}, pages 196--206, 2019.

\bibitem{Thomas2020InvestigatingTI}
A.~Thomas, D.~Adelani, A.~Davody, Aditya Mogadala, and D.~Klakow.
\newblock Investigating the impact of pre-trained word embeddings on
  memorization in neural networks.
\newblock In {\em TDS}, 2020.

\bibitem{brown2020language}
Tom~B Brown, Benjamin Mann, Nick Ryder, Melanie Subbiah, Jared Kaplan, Prafulla
  Dhariwal, Arvind Neelakantan, Pranav Shyam, Girish Sastry, Amanda Askell,
  et~al.
\newblock Language models are few-shot learners.
\newblock {\em arXiv preprint arXiv:2005.14165}, 2020.

\bibitem{raffel2019exploring}
Colin Raffel, Noam Shazeer, Adam Roberts, Katherine Lee, Sharan Narang, Michael
  Matena, Yanqi Zhou, Wei Li, and Peter~J Liu.
\newblock Exploring the limits of transfer learning with a unified text-to-text
  transformer.
\newblock {\em arXiv preprint arXiv:1910.10683}, 2019.

\bibitem{melis2019exploiting}
Luca Melis, Congzheng Song, Emiliano De~Cristofaro, and Vitaly Shmatikov.
\newblock Exploiting unintended feature leakage in collaborative learning.
\newblock In {\em 2019 IEEE Symposium on Security and Privacy (SP)}, pages
  691--706. IEEE, 2019.

\bibitem{song2020information}
Congzheng Song and Ananth Raghunathan.
\newblock Information leakage in embedding models.
\newblock In {\em Proceedings of the 2020 ACM SIGSAC Conference on Computer and
  Communications Security}, pages 377--390, 2020.

\bibitem{pan2020privacy}
Xudong Pan, Mi~Zhang, Shouling Ji, and Min Yang.
\newblock Privacy risks of general-purpose language models.
\newblock In {\em 2020 IEEE Symposium on Security and Privacy (SP)}, pages
  1314--1331. IEEE, 2020.

\bibitem{krishna2019thieves}
Kalpesh Krishna, Gaurav~Singh Tomar, Ankur~P Parikh, Nicolas Papernot, and
  Mohit Iyyer.
\newblock Thieves on sesame street! model extraction of bert-based apis.
\newblock {\em arXiv preprint arXiv:1910.12366}, 2019.

\bibitem{he2020deberta}
Pengcheng He, Xiaodong Liu, Jianfeng Gao, and Weizhu Chen.
\newblock Deberta: Decoding-enhanced bert with disentangled attention.
\newblock {\em arXiv preprint arXiv:2006.03654}, 2020.

\bibitem{inan2016tying}
Hakan Inan, Khashayar Khosravi, and Richard Socher.
\newblock Tying word vectors and word classifiers: A loss framework for
  language modeling.
\newblock {\em arXiv preprint arXiv:1611.01462}, 2016.

\bibitem{press2016using}
Ofir Press and Lior Wolf.
\newblock Using the output embedding to improve language models.
\newblock {\em arXiv preprint arXiv:1608.05859}, 2016.

\bibitem{Loper2002}
Edward Loper and Steven Bird.
\newblock {NLTK:} the natural language toolkit.
\newblock {\em CoRR}, cs.CL/0205028, 2002.

\bibitem{Wolf2020}
Thomas Wolf, Lysandre Debut, Victor Sanh, Julien Chaumond, Clement Delangue,
  Anthony Moi, Pierric Cistac, Tim Rault, R{\'{e}}mi Louf, Morgan Funtowicz,
  Joe Davison, Sam Shleifer, Patrick von Platen, Clara Ma, Yacine Jernite,
  Julien Plu, Canwen Xu, Teven~Le Scao, Sylvain Gugger, Mariama Drame, Quentin
  Lhoest, and Alexander~M. Rush.
\newblock Transformers: State-of-the-art natural language processing.
\newblock In Qun Liu and David Schlangen, editors, {\em Proceedings of the 2020
  Conference on Empirical Methods in Natural Language Processing: System
  Demonstrations, {EMNLP} 2020 - Demos, Online, November 16-20, 2020}, pages
  38--45. Association for Computational Linguistics, 2020.

\bibitem{Papineni2002}
Kishore Papineni, Salim Roukos, Todd Ward, and Wei{-}Jing Zhu.
\newblock Bleu: a method for automatic evaluation of machine translation.
\newblock In {\em Proceedings of the 40th Annual Meeting of the Association for
  Computational Linguistics, July 6-12, 2002, Philadelphia, PA, {USA}}, pages
  311--318. {ACL}, 2002.

\bibitem{rowling2000goblet}
J.~K. Rowling.
\newblock {\em Harry Potter and the Goblet of Fire}.
\newblock Scholastic, United States, 2000.

\bibitem{rowling2006princce}
J.~K. Rowling.
\newblock {\em Harry Potter and the Half-Blood Prince}.
\newblock Scholastic, United States, 2006.

\end{thebibliography}
\bibliographystyle{unsrt}

\appendix
\section{Appendix}
\label{sec:appendix:tables}
\begin{table*}[ht]
    \centering
    \begin{adjustbox}{angle=90, max width=0.6\textwidth}
    \footnotesize
    \begin{tabular}{| l || l l l l || l l l l || l l l l || l l l l |}
        \hline
        %
        % meta header
        & & \multicolumn{3}{c||}{\thead{Precision}}
        & & \multicolumn{3}{c||}{\thead{Precision}}
        & & \multicolumn{3}{c||}{\thead{Precision}}
        & & \multicolumn{3}{c|}{\thead{Precision}}      \\
        %
        % header
        \thead{Dataset}     &
        \thead{BLEU}        &
        \thead{3-gram}      &
        \thead{4-gram}      &
        \thead{Sent}        &
        \thead{BLEU}        &
        \thead{3-gram}      &
        \thead{4-gram}      &
        \thead{Sent}        &
        \thead{BLEU}        &
        \thead{3-gram}      &
        \thead{4-gram}      &
        \thead{Sent}        &
        \thead{BLEU}        &
        \thead{3-gram}      &
        \thead{4-gram}      &
        \thead{Sent}        \\
        %
        % action:
        \hline\hline &
        \multicolumn{4}{r||}{\thead{action$_{\text{100\%: 5,903,010 lines}}$}}  &
        \multicolumn{4}{c||}{\thead{action$_{\text{10\%: 590,818 lines}}$}}     &
        \multicolumn{4}{c||}{\thead{action$_{\text{1\%: 59,330 lines}}$}}       &
        \multicolumn{4}{c}{\thead{action$_{\text{0.1\%: 5,864 lines}}$}}        \\
        \hline
        Action
            & 0.9989        & 0.9986        & 0.9982        & 0.9931
            & 0.9961        & 0.9954        & 0.9936        & 0.9769
            & 0.9566        & 0.9479        & 0.9298        & 0.7899
            & 0.6380        & 0.5794        & 0.4844        & 0.2182    \\
        Drama
            & 0.9982        & 0.9979        & 0.9971        & 0.9889
            & 0.9945        & 0.9934        & 0.9910        & 0.9668
            & 0.9516        & 0.9418        & 0.9222        & 0.7702
            & 0.6404        & 0.5820        & 0.4880        & 0.2211    \\
        Fluff
            & 0.9982        & 0.9978        & 0.9970        & 0.9889
            & 0.9943        & 0.9932        & 0.9908        & 0.9668
            & 0.9541        & 0.9448        & 0.9262        & 0.7776
            & 0.6419        & 0.5839        & 0.4899        & 0.2203    \\
        Gutenberg
            & 0.9758        & 0.9709        & 0.9615        & 0.8693
            & 0.9585        & 0.9502        & 0.9345        & 0.7842
            & 0.8509        & 0.8229        & 0.7732        & 0.4430
            & 0.4834        & 0.4141        & 0.3156        & 0.0945    \\
        %
        % drama
        \hline\hline &
        \multicolumn{4}{r||}{\thead{drama$_{\text{100\%: 4,854,969 lines}}$}}   &
        \multicolumn{4}{c||}{\thead{drama$_{\text{10\%: 484,128 lines}}$}}      &
        \multicolumn{4}{c||}{\thead{drama$_{\text{1\%: 48,569 lines}}$}}        &
        \multicolumn{4}{c}{\thead{drama$_{\text{0.1\%: 4,796 lines}}$}}         \\
        \hline
        Action
            & 0.9981        & 0.9977        & 0.9977        & 0.9884
            & 0.9933        & 0.9920        & 0.9891        & 0.9601
            & 0.9334        & 0.9201        & 0.8938        & 0.7041
            & 0.5603        & 0.4939        & 0.3935        & 0.1641    \\
        Drama
            & 0.9989        & 0.9986        & 0.9986        & 0.9931
            & 0.9954        & 0.9944        & 0.9924        & 0.9723
            & 0.9449        & 0.9339        & 0.9115        & 0.7507
            & 0.5839        & 0.5193        & 0.4202        & 0.1825    \\
        Fluff
            & 0.9981        & 0.9977        & 0.9977        & 0.9884
            & 0.9936        & 0.9922        & 0.9895        & 0.9619
            & 0.9407        & 0.9288        & 0.9054        & 0.7289
            & 0.5812        & 0.5170        & 0.4178        & 0.1767    \\
        Gutenberg
            & 0.9763        & 0.9716        & 0.9716        & 0.8725
            & 0.9563        & 0.9476        & 0.9310        & 0.7710
            & 0.8273        & 0.7956        & 0.7395        & 0.3986
            & 0.4255        & 0.3535        & 0.2566        & 0.0739    \\
        %
        % fluff
        \hline\hline &
        \multicolumn{4}{r||}{\thead{fluff$_{\text{100\%: 5,251,248 lines}}$}}   &
        \multicolumn{4}{c||}{\thead{fluff$_{\text{10\%: 524,322 lines}}$}}      &
        \multicolumn{4}{c||}{\thead{fluff$_{\text{1\%: 52,696 lines}}$}}        &
        \multicolumn{4}{c}{\thead{fluff$_{\text{0.1\%: 5,226 lines}}$}}         \\
        \hline
        Action
            & 0.9972        & 0.9966        & 0.9954        & 0.9827
            & 0.9912        & 0.9894        & 0.9856        & 0.9480
            & 0.9280        & 0.9137        & 0.8853        & 0.6876
            & 0.5841        & 0.5200        & 0.4211        & 0.1810    \\
        Drama
            & 0.9973        & 0.9967        & 0.9956        & 0.9831
            & 0.9914        & 0.9896        & 0.9859        & 0.9491
            & 0.9328        & 0.9193        & 0.8928        & 0.7061
            & 0.6025        & 0.5400        & 0.4425        & 0.1949    \\
        Fluff
            & 0.9988        & 0.9986        & 0.9981        & 0.9928
            & 0.9958        & 0.9949        & 0.9930        & 0.9746
            & 0.9518        & 0.9421        & 0.9224        & 0.7756
            & 0.6306        & 0.5713        & 0.4755        & 0.2172    \\
        Gutenberg
            & 0.9726        & 0.9671        & 0.9564        & 0.8453
            & 0.9495        & 0.9393        & 0.9201        & 0.7305
            & 0.8126        & 0.7784        & 0.7186        & 0.3722
            & 0.4415        & 0.3701        & 0.2724        & 0.0804    \\
        %
        % gutenberg
        \hline\hline &
        \multicolumn{4}{r||}{\thead{gutenberg$_{\text{100\%: 2,728,188 lines}}$}}   &
        \multicolumn{4}{c||}{\thead{gutenberg$_{\text{10\%: 273,263 lines}}$}}      &
        \multicolumn{4}{c||}{\thead{gutenberg$_{\text{1\%: 27,240 lines}}$}}        &
        \multicolumn{4}{c}{\thead{gutenberg$_{\text{0.1\%: 2,755 lines}}$}}         \\
        \hline
        Action
            & 0.9833        & 0.9798        & 0.9728        & 0.9007
            & 0.9651        & 0.9579        & 0.9438        & 0.8104
            & 0.8460        & 0.8163        & 0.7623        & 0.4446
            & 0.3440        & 0.2685        & 0.1707        & 0.0496    \\
        Drama
            & 0.9846        & 0.9814        & 0.9750        & 0.9089
            & 0.9676        & 0.9608        & 0.9477        & 0.8223
            & 0.8561        & 0.8280        & 0.7773        & 0.4686
            & 0.3551        & 0.2793        & 0.1799        & 0.0525    \\
        Fluff
            & 0.9806        & 0.9766        & 0.9687        & 0.8878
            & 0.9616        & 0.9536        & 0.9383        & 0.7947
            & 0.8451        & 0.8152        & 0.7618        & 0.4433
            & 0.3510        & 0.2759        & 0.1765        & 0.0503    \\
        Gutenberg
            & 0.9972        & 0.9966        & 0.9955        & 0.9744
            & 0.9894        & 0.9873        & 0.9830        & 0.9110
            & 0.8933        & 0.8727        & 0.8346        & 0.5041
            & 0.3362        & 0.2627        & 0.1693        & 0.0421    \\
        \hline
    \end{tabular}
    \end{adjustbox}
    \caption{InvBERT Linear trained on every data sizes and evaluated across all eval datasets.}
    \label{fig:table:linear}
\end{table*}
\begin{table*}[ht]
    \centering
    \begin{adjustbox}{angle=90, max width=0.6\textwidth}
    \footnotesize
    \begin{tabular}{| l || l l l l || l l l l || l l l l || l l l l |}
        \hline
        %
        % meta header
        & & \multicolumn{3}{c||}{\thead{Precision}}
        & & \multicolumn{3}{c||}{\thead{Precision}}
        & & \multicolumn{3}{c||}{\thead{Precision}}
        & & \multicolumn{3}{c|}{\thead{Precision}}      \\
        %
        % header
        \thead{Dataset}     &
        \thead{BLEU}        &
        \thead{3-gram}      &
        \thead{4-gram}      &
        \thead{Sent}        &
        \thead{BLEU}        &
        \thead{3-gram}      &
        \thead{4-gram}      &
        \thead{Sent}        &
        \thead{BLEU}        &
        \thead{3-gram}      &
        \thead{4-gram}      &
        \thead{Sent}        &
        \thead{BLEU}        &
        \thead{3-gram}      &
        \thead{4-gram}      &
        \thead{Sent}        \\
        %
        % action:
        \hline\hline &
        \multicolumn{4}{r||}{\thead{action$_{\text{100\%: 5,903,010 lines}}$}}  &
        \multicolumn{4}{c||}{\thead{action$_{\text{10\%: 590,818 lines}}$}}     &
        \multicolumn{4}{c||}{\thead{action$_{\text{1\%: 59,330 lines}}$}}       &
        \multicolumn{4}{c}{\thead{action$_{\text{0.1\%: 5,864 lines}}$}}        \\
        \hline
        Action
            & 0.9797        & 0.9762        & 0.9692        & 0.9174
            & 0.9443        & 0.9376        & 0.9190        & 0.8049
            & 0.6792        & 0.6360        & 0.5522        & 0.4009
            & 0.1253        & 0.0737        & 0.0338        & 0.0251    \\
        Drama
            & 0.9675        & 0.9625        & 0.9521        & 0.8722
            & 0.9290        & 0.9222        & 0.9000        & 0.7611
            & 0.6679        & 0.6263        & 0.5418        & 0.3496
            & 0.1274        & 0.0762        & 0.0352        & 0.0254    \\
        Fluff
            & 0.9632        & 0.9579        & 0.9468        & 0.8625
            & 0.9252        & 0.9186        & 0.8961        & 0.7547
            & 0.6656        & 0.6241        & 0.5397        & 0.3468
            & 0.1266        & 0.0761        & 0.0353        & 0.0256    \\
        Gutenberg
            & 0.8299        & 0.8048        & 0.7664        & 0.5252
            & 0.7284        & 0.6919        & 0.6358        & 0.3930
            & 0.4009        & 0.3515        & 0.2631        & 0.1430
            & 0.0608        & 0.0354        & 0.0114        & 0.0117    \\
        %
        % drama
        \hline\hline &
        \multicolumn{4}{r||}{\thead{drama$_{\text{100\%: 4,854,969 lines}}$}}   &
        \multicolumn{4}{c||}{\thead{drama$_{\text{10\%: 484,128 lines}}$}}      &
        \multicolumn{4}{c||}{\thead{drama$_{\text{1\%: 48,569 lines}}$}}        &
        \multicolumn{4}{c}{\thead{drama$_{\text{0.1\%: 4,796 lines}}$}}         \\
        \hline
        Action
            & 0.9581        & 0.9508        & 0.9376        & 0.8356
            & 0.9084        & 0.8934        & 0.8656        & 0.7058
            & 0.5964        & 0.5513        & 0.4581        & 0.2789
            & 0.1094        & 0.0628        & 0.0279        & 0.0195    \\
        Drama
            & 0.9760        & 0.9719        & 0.9636        & 0.9050
            & 0.9331        & 0.9219        & 0.8999        & 0.7813
            & 0.6173        & 0.5689        & 0.4769        & 0.3093
            & 0.1140        & 0.0658        & 0.0295        & 0.0206    \\
        Fluff
            & 0.9589        & 0.9521        & 0.9396        & 0.8470
            & 0.9122        & 0.8979        & 0.8715        & 0.7216
            & 0.6033        & 0.5579        & 0.4654        & 0.2931
            & 0.1129        & 0.0653        & 0.0295        & 0.0204    \\
        Gutenberg
            & 0.8189        & 0.7928        & 0.7523        & 0.5084
            & 0.6998        & 0.6610        & 0.6018        & 0.3696
            & 0.3492        & 0.3036        & 0.2166        & 0.1206
            & 0.0538        & 0.0266        & 0.0083        & 0.0096    \\
        %
        % fluff
        \hline\hline &
        \multicolumn{4}{r||}{\thead{fluff$_{\text{100\%: 5,251,248 lines}}$}}   &
        \multicolumn{4}{c||}{\thead{fluff$_{\text{10\%: 524,322 lines}}$}}      &
        \multicolumn{4}{c||}{\thead{fluff$_{\text{1\%: 52,696 lines}}$}}        &
        \multicolumn{4}{c}{\thead{fluff$_{\text{0.1\%: 5,226 lines}}$}}         \\
        \hline
        Action
            & 0.9501        & 0.9416        & 0.9262        & 0.8117
            & 0.9050        & 0.8894        & 0.6104        & 0.6926
            & 0.6079        & 0.5661        & 0.4757        & 0.2856
            & 0.1161        & 0.0705        & 0.0325        & 0.0227    \\
        Drama
            & 0.9551        & 0.9475        & 0.9333        & 0.8315
            & 0.9112        & 0.8894        & 0.6104        & 0.7156
            & 0.6185        & 0.5744        & 0.4851        & 0.3029
            & 0.1204        & 0.0737        & 0.0343        & 0.0234    \\
        Fluff
            & 0.9751        & 0.9709        & 0.9626        & 0.9064
            & 0.9369        & 0.9265        & 0.6104        & 0.8007
            & 0.6499        & 0.6013        & 0.5147        & 0.3416
            & 0.1239        & 0.0749        & 0.0353        & 0.0243    \\
        Gutenberg
            & 0.8015        & 0.7731        & 0.7289        & 0.4706
            & 0.6502        & 0.6104        & 0.5499        & 0.3499
            & 0.3497        & 0.3077        & 0.2215        & 0.1223
            & 0.0541        & 0.0287        & 0.0091        & 0.0098    \\
        %
        % gutenberg
        \hline\hline &
        \multicolumn{4}{r||}{\thead{gutenberg$_{\text{100\%: 2,728,188 lines}}$}}   &
        \multicolumn{4}{c||}{\thead{gutenberg$_{\text{10\%: 273,263 lines}}$}}      &
        \multicolumn{4}{c||}{\thead{gutenberg$_{\text{1\%: 27,240 lines}}$}}        &
        \multicolumn{4}{c}{\thead{gutenberg$_{\text{0.1\%: 2,755 lines}}$}}         \\
        \hline
        Action
            & 0.9028        & 0.8904        & 0.8619        & 0.6513
            & 0.8051        & 0.7848        & 0.7313        & 0.4522
            & 0.3474        & 0.2736        & 0.1852        & 0.0915
            & 0.0397        & 0.0161        & 0.0037        & 0.0049    \\
        Drama
            & 0.9124        & 0.9019        & 0.8761        & 0.6813
            & 0.8164        & 0.7976        & 0.7461        & 0.4766
            & 0.3546        & 0.2809        & 0.1916        & 0.0975
            & 0.0417        & 0.0172        & 0.0042        & 0.0050    \\
        Fluff
            & 0.8964        & 0.8838        & 0.8545        & 0.6419
            & 0.7993        & 0.7784        & 0.7246        & 0.4494
            & 0.3450        & 0.2718        & 0.1836        & 0.0929
            & 0.0396        & 0.0161        & 0.0038        & 0.0044    \\
        Gutenberg
            & 0.9330        & 0.9241        & 0.9071        & 0.7528
            & 0.8094        & 0.7839        & 0.7419        & 0.4917
            & 0.2803        & 0.2175        & 0.1406        & 0.0764
            & 0.0314        & 0.0141        & 0.0032        & 0.0070    \\
        \hline
    \end{tabular}
    \end{adjustbox}
    \caption{InvBERT Seq2Seq trained on every data size and evaluated across all eval datasets.}
    \label{fig:table:seq2seq}
\end{table*}

\end{document}